\pdfoutput=1

\pdfoutput=1

\documentclass[11pt]{article}

\usepackage[]{acl}

\usepackage{times}
\usepackage{latexsym}
\usepackage[capitalize,noabbrev]{cleveref}
\usepackage{afterpage}
\usepackage{adjustbox}
\usepackage{makecell}
\usepackage{array}

\usepackage[T1]{fontenc}

\usepackage[utf8]{inputenc}

\usepackage{microtype}

\usepackage{inconsolata}

\usepackage{graphicx}

\usepackage{float}

\usepackage{xcolor}

\usepackage{tcolorbox}
\usepackage{siunitx}
\usepackage{caption}
\usepackage{booktabs}
\usepackage{makecell}
\usepackage{graphicx}
\usepackage{hyperref}
\usepackage{booktabs}
\usepackage{wrapfig}

\usepackage{tabularx}

\usepackage{listings}
\usepackage{booktabs,multirow,array,graphicx,pifont}

\usepackage{xcolor} 
\usepackage{ifthen}
\newboolean{showcomments}
\setboolean{showcomments}{true} 

\title{Left, Right, or Center? Evaluating LLM Framing in News Classification and Generation}

\author{
  Molly Kennedy\textsuperscript{1}, Ali Parker\textsuperscript{1}, Yihong Liu\textsuperscript{1}, Hinrich Schütze\textsuperscript{1} \\
  \textsuperscript{1}Ludwig-Maximilians-Universität München \\
  \texttt{molly@cis.uni-muenchen.de} \\
  \texttt{}
}

\newcounter{notecounter}

\newcommand{\enoteson}{\long\gdef\enote##1##2{{
\stepcounter{notecounter}
{\large\bf
\hspace{0cm}\arabic{notecounter} $<<<$ ##1: ##2
$>>>$\hspace{1cm}}}}}

\enoteson

\begin{document}
\maketitle

\begin{abstract}
Large Language Model (LLM) based summarization and text generation are increasingly used for producing and rewriting text, raising concerns about political framing in journalism where subtle wording choices can shape interpretation.
Across nine state-of-the-art LLMs, we study political framing by testing whether LLMs' \emph{classification-based} bias signals align with framing behavior in their \emph{generated} summaries.
We first compare few-shot ideology predictions against \textsc{Left/Center/Right} labels. We then generate ``steered'' summaries under \textsc{faithful}, \textsc{centrist}, \textsc{left}, and \textsc{right} prompts, and score all outputs using a single fixed ideology evaluator. 
We find pervasive ideological center-collapse in both article-level ratings and generated text, indicating a systematic tendency toward centrist framing. 
Among evaluated models, Grok 4 is by far the most ideologically expressive generator, while Claude Sonnet 4.5 and Llama 3.1 achieve the strongest bias-rating performance among commercial and open-weight models, respectively.
\end{abstract}

\section{Introduction}
\label{sec:intro}

Media bias is often expressed through \emph{ideological framing}: outlets can report on the same event while foregrounding different values, priorities, and causal narratives, shaping how readers interpret political and economic issues \citep{mokhberian2020moral,pastorino2024decoding}. Such framing is frequently subtle and context-dependent, and even human judgments of bias can vary across annotators and label schemes \citep{spinde2023media}. As large language models (LLMs) are increasingly integrated into writing workflows for summarization and rewriting, understanding how they handle framing in news becomes a practical concern \citep{bavaresco-etal-2025-llms, wang2025media}. Furthermore, masses of short-form AI generated misinformation have taken social media channels by storm \citep{zhou2023synthetic}, underscoring the need to better understand how easily models can be manipulated for ideological framing.

Most work on LLM political behavior evaluates models in \emph{classification}-like settings (e.g., predicting ideology labels or answering political questionnaires) \citep{rottger2024political, elbouanani2025analyzing, haller2025leveraging}.
Building on this paradigm, more recent work has begun to evaluate prompted LLMs for media-bias detection across model families \citep{maab2024media, faulborn2025only}.
By contrast, many real deployments are \emph{generative}: models produce headlines and summaries that may introduce shifts in emphasis or tone even when factual content is preserved. 

\begin{figure}[t!]
\includegraphics[width=0.5\textwidth]{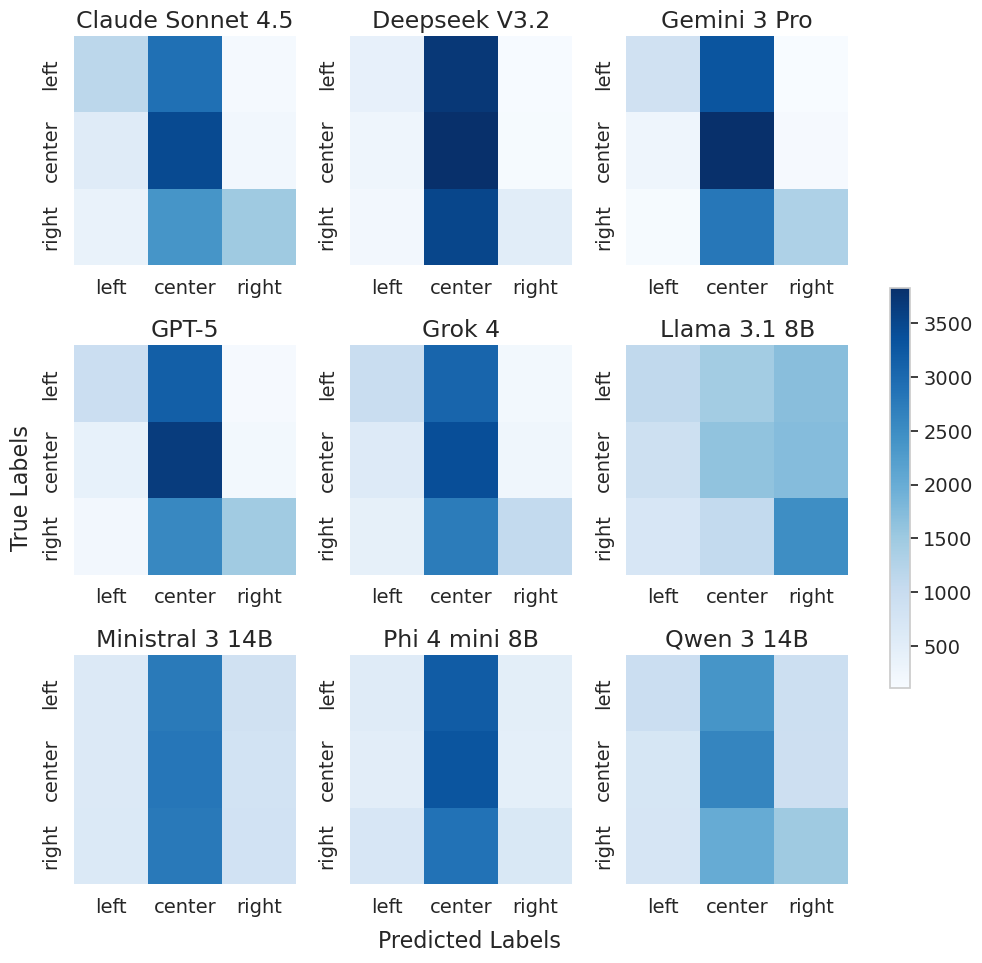}
\caption{Confusion matrices for LLM political direction rating capability on 12k (balanced) political news articles.}
\label{fig:conf}
\end{figure}
Evidence on generative media bias and perspective drift is emerging, but remains more limited than classification-focused analyses, especially for realistic news inputs and for comparisons across open and commercial systems \citep{liu2024p3sum,trhlik2024quantifying}. This motivates two questions: (1) do models exhibit systematic ideological tendencies when asked to \emph{label} news (see \cref{fig:conf}), and (2) is their \emph{generated} framing consistent with these labeling behaviors? (see \cref{tab:res_gen}).

To address these questions, we evaluate political framing across nine state-of-the-art LLMs spanning both commercial and open-weight variants.
Rather than asserting a single ground-truth ideology for each article, our goal is to characterize model behavior relative to an external reference labeling scheme and under controlled prompting conditions.
Using the AllSides \textsc{Left/Center/Right} label space \citep{baly2020we}, we evaluate two complementary behaviors: \textbf{label alignment}, measured via few-shot bias classification, and \textbf{generation behavior}, measured via perspective-conditioned summary generation. 
We then compare these two modes to test whether models that appear aligned or ``centrist'' in classification maintain similar framing when generating user-facing news text.

Our contributions are threefold:
(i) we provide a comparative analysis of open and commercial LLMs on AllSides-aligned political bias classification;
(ii) we introduce a controlled evaluation of perspective prompts for summary generation, including diagnostics for centrist defaulting; and 
(iii) we conduct a lexical framing analysis that links prompt-induced stylistic changes to measured ideological shifts \citep{monroe2008fightin}. 

\section{Related Work}
\label{sec:related}

\paragraph{Media bias and ideology in news.}
Detecting political ideology and media bias from news text has a long history in NLP. Early and modern approaches use article content, headlines, and metadata to predict \textsc{left/center/right} ideology, and emphasize challenges such as domain shift and source confounding \citep{kulkarni2018multi, baly2020we}. Datasets such as BASIL provide finer-grained annotations that distinguish lexical bias from informational bias, reflecting how framing can emerge through selection and emphasis rather than overtly partisan language \citep{fan2019plain}. Surveys highlight that media bias is multifaceted (e.g., framing, gatekeeping, tone), complicating evaluation when reduced to a single label \citep{spinde2023media}. Our work uses coarse \textsc{Left/Center/Right} labels as a pragmatic reference to elucidate model behaviors related to agreement, skewness, and center-collapse.

\paragraph{Political bias and framing in LLMs.}
Recent studies examine political bias in LLMs across both \emph{content} (what is stated) and \emph{style} (how it is framed), showing systematic differences across models and prompting setups \citep{bang2024measuring}. Other work moves toward journalism-like settings by analyzing bias in generated news content and how it differs from human writing \citep{trhlik2024quantifying}. Related research in summarization notes that preserving author perspective or political stance is non-trivial and can drift under standard objectives, motivating perspective-preserving methods \citep{liu2024p3sum}. 

\paragraph{Steerability and controllable generation.}
Prompting is a primary mechanism for controlling LLM outputs. Recent work proposes benchmarks and metrics for steerability, finding asymmetries and limits in how reliably prompts change behavior \citep{miehling2025evaluating}. Controllable generation spans prompt engineering, decoding-time controls, and model-based interventions \citep{liu2024p3sum}. Our approach is lightweight and model-agnostic: by using a single fixed evaluator to score all generated outputs, we obtain comparable steering-strength estimates and diagnose \emph{center-defaulting}, a practical failure mode in perspective-conditioned news generation.  To obtain comparable measurements across models and conditions, we score generated outputs with a single fixed LLM evaluator. Prior work shows that LLM judges can exhibit systematic biases, motivating careful prompt design \citep{chen2024humans}.

\section{Dataset}
\label{sec:data}

The AllSides news-ideology corpus of \citet{baly-etal-2020-detect} is used. The corpus is composed of $\sim$35k news articles labeled with a coarse ideology label in \textsc{Left, Center, Right}.
Each instance includes the article title and extracted body text (plus metadata such as source/outlet and URL). 
We use the title as the \emph{headline} input and the body text as the main \emph{article} input. 
A balanced subset of the corpus is used: for Stage~1 (bias classification), a sample of 12k articles is stratified to balance \textsc{Left, Center, Right} labels. 
For Stage~2 (summary generation), we use a separate balanced subset of 1k articles, again stratified by label.

\section{Methodology}
\label{sec:method}

We evaluate political framing behavior of nine LLMs in two stages using the AllSides \textsc{Left/Center/Right} label space.

\paragraph{Data.}
Stage~1 employs a 12k stratified subset (balanced across labels) for bias classification. Stage~2 uses a separate 1k balanced subset for generation. 

\paragraph{Stage 1: Political rating alignment analysis via Bias Classification}
Each model predicts a single label in \{\textsc{Left, Center, Right}\} via a fixed few-shot prompt applied to the title and body text (few-shot examples held constant across models). 
Table~\ref{tab:prompts_overview} summarizes our prompt templates; full templates are provided in Appendix~\ref{app:prompts}. Deterministic decoding is enabled when available (e.g., greedy decoding / temperature $=0$). We report accuracy, macro-F1, Cohen's $\kappa$, confusion matrices, and distributional diagnostics including prediction skew and \emph{center-collapse} (over-predicting \textsc{Center}).

\begin{table}[hbt!]
\centering
\footnotesize
\setlength{\tabcolsep}{3pt}
\renewcommand{\arraystretch}{1.08}

\begin{adjustbox}{max width=0.92\columnwidth} 
\begin{tabularx}{\columnwidth}{@{}l l X@{}}
\toprule
\textbf{Prompt} & \textbf{Input} & \textbf{Output} \\
\midrule

\multicolumn{3}{@{}l@{}}{\textbf{Stage 1}} \\
\hspace{1em}Classification & title + text &
\parbox[t]{\linewidth}{\raggedright ideology label\\(Left/Center/Right)} \\

\multicolumn{3}{@{}l@{}}{\textbf{Stage 2}} \\
\hspace{1em}Summary & title + text &
\parbox[t]{\linewidth}{\raggedright perspective-conditioned\\\textasciitilde100-token summary} \\

\multicolumn{3}{@{}l@{}}{\textbf{Evaluator}} \\
\hspace{1em}Ideology labeling & generated text &
\parbox[t]{\linewidth}{\raggedright ideology label\\(Left/Center/Right)} \\

\bottomrule
\end{tabularx}
\end{adjustbox}

\caption{Prompt families used in our experiments. Full templates are in Appendix~\ref{app:prompts}.}
\label{tab:prompts_overview}
\end{table}
\paragraph{Stage 2: Alignment under ideologically steered summary generation}
For each item, we generate a one-line $\sim$100-token summary under four prompt conditions: \textsc{faithful}, \textsc{centrist}, \textsc{left}, and \textsc{right}. Prompts include three fixed few-shot examples illustrating \textsc{Left/Center/Right} framing;
any prefixed fields are stripped in post-processing. For open-weight models run via \texttt{transformers}, greedy decoding (\texttt{do\_sample=False}, \texttt{temperature=0}) is used. 
We target comparable output lengths via strict formatting and length instructions; maximum generation limits are set per model 
and are held fixed across prompt conditions within each model.

\paragraph{Ideology evaluation.}
We score each generated output with a single fixed ideology evaluator, \emph{Gemini 3 Pro}, using a fixed prompt with the same label set and three few-shot examples. The evaluator is instructed to use only the provided text and output exactly one label in the format ``Label: \{\textsc{left}$\mid$\textsc{center}$\mid$\textsc{right}\}''; inputs are truncated to 12{,}000 characters for cost/control. We report label distributions, prompt-induced label shifts relative to the source label, and \emph{center-defaulting} (\textsc{Left/Right}-prompt outputs labeled \textsc{Center}).

\section{Experiment Results}
\label{sec:results}

\paragraph{Stage 1: political rating alignment performance} (Table \ref{tab:res_rating}) is led by Claude Sonnet 4.5, with GPT-5 following closely. Llama 3.1 and Qwen 3 are comfortably the best raters amongst the open-weight models tested, with performance competitive with the closed-source models. Rating performance across the board leaves significant room for improvement. 

\begin{table}[hbt!]
\centering
\small
\setlength{\tabcolsep}{3.5pt} 
\begin{adjustbox}{max width=\columnwidth}
\begin{tabular}{@{}lccc@{}}
\toprule
Model & Macro-F1 & Acc & $\kappa$ \\
\midrule
Claude Sonnet 4.5 & \textbf{0.462} & \textbf{0.480} & \textbf{0.221} \\
Deepseek V3.2     & 0.289          & 0.373          & 0.060          \\
Gemini 3 Pro      & 0.433          & 0.470          & 0.205          \\
GPT-5             & 0.448          & 0.475          & 0.213          \\
Grok 4            & 0.398          & 0.429          & 0.143          \\
\midrule
Llama 3.1 8B      & \textbf{0.397} & \textbf{0.408} & \textbf{0.112} \\
Ministral 3 14B   & 0.299          & 0.335          & 0.003          \\
Phi 4 mini 8B     & 0.299          & 0.354          & 0.031          \\
Qwen 3 14B        & 0.380          & 0.396          & 0.094          \\
\bottomrule
\end{tabular}
\end{adjustbox}
\caption{Model rating performance on 12000 (balanced) political news articles.}
\label{tab:res_rating}
\end{table}

\paragraph{Granular insight into the models' bias rating calibration} is displayed in Figure \ref{fig:conf}. An important condition for rating performance is a model's tendency to avoid central collapse. Calibration for more confident \textsc{left} and \textsc{right} ratings is beneficial. All of the closed-source models exhibit prominent biases towards centrist ratings. In line with Table \ref{tab:res_rating}, top performers Claude Sonnet 4.5 and GPT-5 tend to rate articles as \textsc{left} and \textsc{right} more frequently than the others.

For open-weight models, this centrist tendency is also the case for Ministral 3, Phi 4 and Qwen 3, but to a lesser extent. Uniquely, LLama 3.1 displays a left-wing bias, with predicted ratings tending further right of ground truth. In line with the performance metrics, Llama 3.1 and Qwen 3 exhibit the highest tendency to rate articles as \textsc{left} or \textsc{right} amongst the open-weight models.



\begin{figure}[hbt!]
\includegraphics[width=0.5\textwidth]{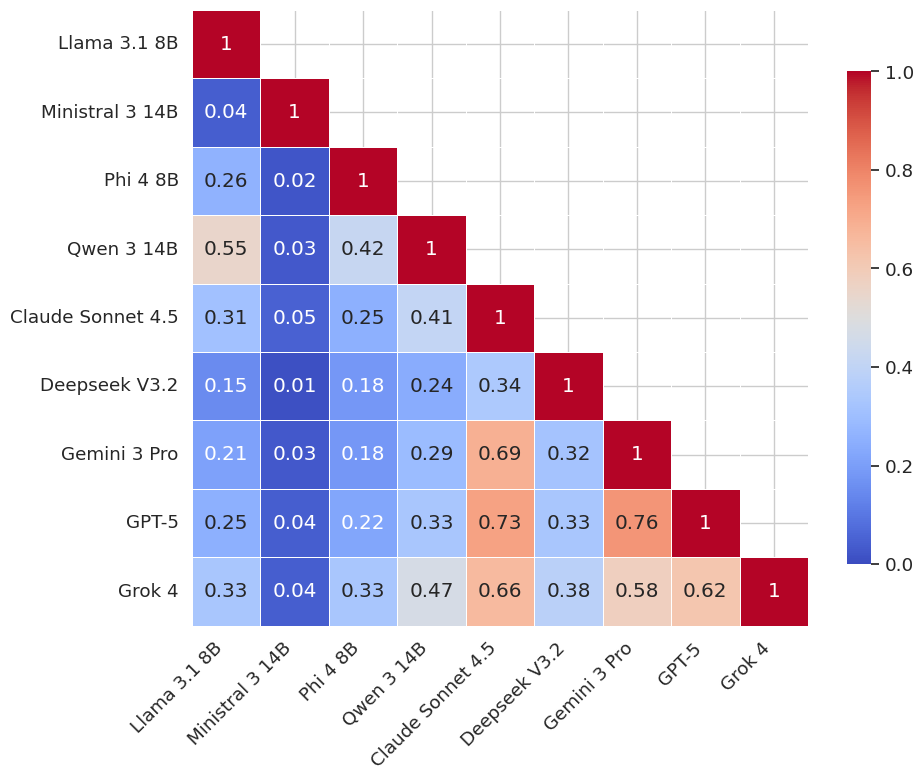}
\caption{Cohen's kappa score heatmap measuring the agreeability between each of the model's ratings as well as with the ground truth.}
\label{fig:cohen_all}
\end{figure}

\paragraph{Inter-model agreement elucidates common alignment across models.} Measured using Cohen's kappa score (Figure \ref{fig:cohen_all}), we find that all of the commercial models with the exception of Deepseek tend agree with each other to a large extent.

Notably, Llama 3.1 and Qwen 3 do not agree with any of the closed models to this extent, despite competitive performance. They also do not agree with each other to a commensurate extent. Comparable performance in these models groups is therefore achieved via (at least slightly) different sets of ratings. Exploring why this is the case would be interesting future work.

\paragraph{For Stage 2: Alignment under ideologically steered summary generation,} our findings are presented in Table \ref{tab:res_rating}. We find that Grok 4 is most effective at both imbuing political ideology into text summaries, and preserving political tone of the original text. The model's ability to generate right-leaning summaries in particular draws a stark contrast to both open-weight and commercial models. In general, left-steered summaries are more accurately evaluated across models than their right-steered counterparts. The mediocre performance for faithful summaries is attributed to the fact that the original articles are not as prominently pointed in a particular direction when compared to summaries with explicit instruction to incorporate bias.

\begin{table}[hbt!]
\centering
\small
\setlength{\tabcolsep}{3.5pt} 
\begin{tabular}{@{}lcccc@{}}
\toprule
Model & Faithful & Left & Center & Right \\
\midrule
Claude Sonnet 4.5 & 0.426 & 0.845 & \textbf{0.974} & 0.613 \\
Deepseek V3.2     & 0.409 & 0.519 & 0.952          & 0.324 \\
Gemini 3 Pro      & 0.435 & 0.644 & 0.952          & 0.527 \\
GPT-5             & 0.431 & 0.888 & 0.936          & 0.605 \\
Grok 4            & \textbf{0.442} & \textbf{0.953} & 0.924 & \textbf{0.834} \\
\midrule
Llama 3.1 8B      & 0.380 & 0.249 & 0.915 & 0.076 \\
Ministral 3 14B   & 0.401 & \textbf{0.724} & \textbf{0.935} & \textbf{0.170} \\
Phi 4 mini 8B     & 0.395 & 0.174 & 0.908 & 0.099 \\
Qwen 3 14B        & \textbf{0.408} & 0.118 & 0.894 & 0.068 \\
\bottomrule
\end{tabular}
\caption{Article summary generation accuracy (1000 balanced samples) rated by Gemini 3 Pro. Closed models in the upper half and open models in the lower half.}
\label{tab:res_gen}
\end{table}

Ministral 3 performs best amongst the open-weight models, with performance in generating left-leaning summaries competitive to the closed models. \textbf{Open-weight models struggled to generate right-leaning summaries} across the board.


Across models, \textbf{most misclassifications of generated outputs collapse to \textsc{Center} (center-defaulting),} and \textsc{left/right} prompts rarely flip to the opposing label. Center-collapse is therefore a unifying failure mode across both stages. 
One possible explanation is conservative safety behavior that favors ``safe'' centrist language: over-conservative safety alignment is known to induce overly cautious behavior such as overrefusal \citep{pan2025understanding}, and safety guardrails can systematically alter (and sometimes degrade) generation behavior \citep{bonaldi2024safer}.  A potential interpretation for Grok's superior ideological expressivity is that it exhibits a less conservative refusal/guardrail profile than other systems; xAI’s model cards emphasize avoiding over-refusal on sensitive or controversial queries, supported by independent safety audits \citep{akiri2025safety}. This contrast between Grok 4 and other systems underscores how safety design choices can materially shape political framing behavior.



\section{Conclusion}
\label{sec:conclusion}

Overall, this work shows that political framing in LLMs is strongly shaped by safety alignment, with center-collapse emerging as a dominant and cross-task failure mode.  Political bias rating ability ranges from poor to moderate across models, with Claude Sonnet 4.5 leading in performance. Llama 3.1 was the most competitive open-weight alternative. Differences between classification and generation across models point to ideological behavior being a controlled design outcome. 
Grok 4 demonstrated the greatest degree of ideological expressivity for steered summary generation, underscoring the importance of the careful tradeoffs between safety and expressivity as LLMs increasingly mediate political information.

\section{Limitations}
\label{sec:limitations}

\paragraph{Our generation results are evaluated with a single fixed LLM judge (Gemini 3 Pro)} rather than new human annotations for the generated headlines and summaries. This enables consistent scoring across models and prompt conditions but may reflect evaluator-specific biases or prompt sensitivity; reported shifts and steering behavior should therefore be interpreted relative to this evaluator. While this inference is largely attributed to the models' generative abilities (steered summaries are more biased than human-written articles), some random variation must be accounted for due to the Gemini 3 Pro evaluator.

\paragraph{Our conclusions also depend on the AllSides \textsc{Left/Center/Right}} taxonomy and the particular few-shot prompts used for classification and generation. These labels are coarse and may not capture fine-grained or issue-specific framing, and alternative label schemes or prompt designs could change absolute agreement and skew statistics.

\paragraph{Finally, we compare both open-weight and commercial models.} Differences in interfaces and controls (e.g., context limits and decoding parameters) may affect outputs despite efforts to standardize settings. We also cannot verify that our prompted Stage~1 setup matches the original annotation guidance used to produce the AllSides labels, so we treat them as a practical reference rather than definitive ground truth.

\clearpage

\bibliography{anthology1,custom}

\appendix

\clearpage

\section{Appendix}
\label{sec:appendix}

\appendix
\section{Prompts}
\label{app:prompts}

\subsection{Stage 2 Summary Prompt}
This prompt is used to generate a short summary of a news sample conditioned on a specified target perspective. The model receives the article heading and full text, along with the placeholder \texttt{\{PERSPECTIVE\}} indicating the desired framing. Output is constrained to a single line in the fixed format \texttt{Summary: <\dots>} to simplify parsing and downstream evaluation.
\begin{lstlisting}
You are given the HEADING and TEXT of a news sample ...
Target PERSPECTIVE: {PERSPECTIVE}

NEW SAMPLE:
Heading:
{HEADING}

Sample text:
{TEXT}

Output format MUST be exactly one line:
Summary: <your summary>
\end{lstlisting}

\subsection{Evaluator Prompt (Gemini 3 Pro)}
This prompt is used as an automatic evaluator to assign a coarse political-leaning label to a given text. The model is provided only the content via the placeholder \texttt{\{TEXT\}} and must choose exactly one label from \texttt{left}, \texttt{center}, or \texttt{right}. Output is restricted to the fixed one-line format \texttt{Label: left|center|right} to ensure consistent, machine-readable annotations.
\begin{lstlisting}
Classify the political leaning as exactly one of: left, center, right ...
Content:
{TEXT}

Output must be exactly:
Label: left|center|right
\end{lstlisting}

\end{document}